\documentclass[10pt,twocolumn,letterpaper]{article}

\usepackage{cvpr}              
\usepackage{graphicx}

\usepackage{mathtools}
\usepackage{makecell}
\usepackage{hyperref}
\usepackage{url}
\usepackage{epsfig}
\usepackage{amsmath}
\usepackage{amssymb}

\def\cvprPaperID{*****} 
\def\confName{CVPR}
\def\confYear{2022}

\begin{document}

\title{kDecay: Just adding k-decay items on Learning-Rate Schedule to improve Neural Networks}








\author{Tao Zhang\\
  {\tt\small atztao@gmail.com}\\
  Central China Normal University
  \and
  Wei Li\\
  {\tt\small liw@mail.ccnu.edu.cn}\\
  Central China Normal University
  
}

\maketitle

\begin{abstract}
 Recent work has shown that optimizing the Learning Rate (LR) schedule can be a very
accurate and efficient way to train deep neural networks. We observe that the
rate of change (ROC) of LR has correlation with the training process, but how to
use  this relationship to control the training to achieve the purpose of
improving accuracy? We propose a new method, k-decay, just add an extra item to
the commonly used and easy LR schedule(exp, cosine and polynomial), is effectively improves the performance of these
schedule, also better than the state-of-the-art algorithms of LR shcedule such
as SGDR, CLR and AutoLRS. In
the k-decay, by adjusting the hyper-parameter \(k\), to generate different LR
schedule, when k increases, the performance is improved. We evaluate the k-decay
method on CIFAR And ImageNet datasets with different neural networks (ResNet,
Wide ResNet). Our experiments show that this method can improve on most of them.
The accuracy has been improved by 1.08\% on the CIFAR-10 dataset and by 2.07 \%
on the CIFAR-100 dataset. On the ImageNet, accuracy is improved by 1.25\%. Our
method is not only a general method to be applied other LR Shcedule, but also
has no additional computational cost.


\end{abstract}
\section{Introduction}
\label{sec:org326e3c1}
Deep learning~\cite{lecun2015deeplearning} is widely used in image recognition, speech recognition, and many
other fields. Now we have convolutional neural networks for
images~\cite{lin2013network,hu2017squeezeandexcitation,journals/corr/DongLHT15,goodfellow2014generative},
recurrent neural networks for speeches~\cite{devlin2018pretraining}, and graph neural
networks for graphs~\cite{wu2019comprehensive,Kipf:2016tc}.
With the development of technology, our research goal is to obtain better model performance under the same resource conditions. To achieve this goal, we studied
the learning rate schedule from a new perspective, and some research~\cite{loshchilov2016sgdr}~\cite{conf/wacv/Smith17}~\cite{conf/wacv/HsuehLW19} shows
a good schedule for the learning rate can improve the training
performance.

In deep neural networks, the parameters are updated by Stochastic
Gradient Descent (SGD). The formula is \(  w^{\prime}=w-\eta \triangledown
\mathcal L\), where \(w\) and $w^{\prime}$ are parameters, \(\eta\) is the learning rate, and \(\mathcal L\) is the loss function. The \(\eta\) controls the update speed of parameters. When \(\eta\) is large, the model converges very quickly but may skip local minimum
values. When \(\eta\) is small, local minimum values can be found, but the model
converges slowly. The LR schedule \(\eta(t)\) governing the decay from maximum
LR \(\eta_0\) to the minimum LR
\(\eta_e\) can solve this contradiction.

\begin{figure}[!t]
\centering
\centerline{\includegraphics[width=\linewidth]{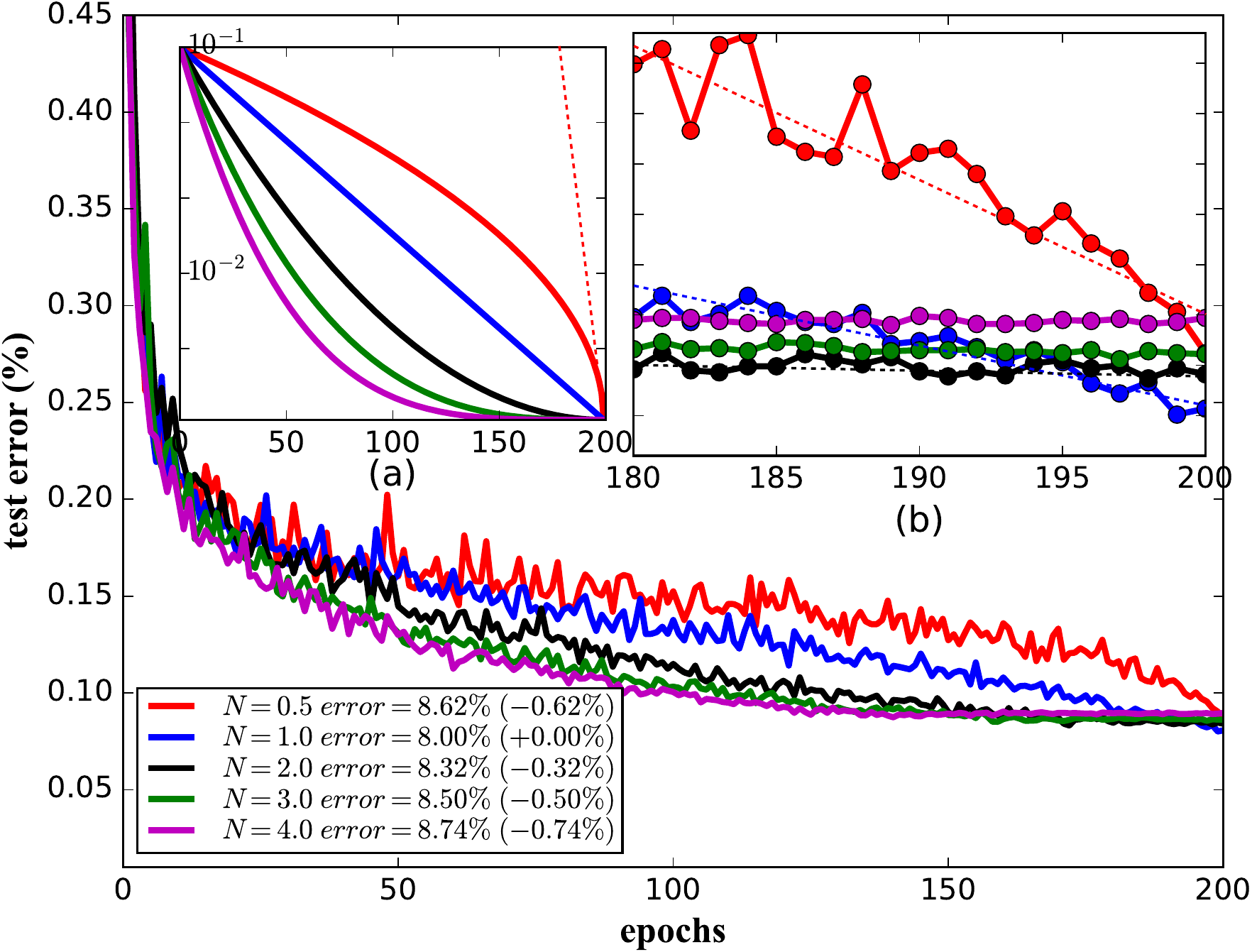}}
\caption{Testing errors on CIFAR-10 with ResNet-20, with different \(N\) on polynomial decay. (a) is the learning rate curve of the polynomial decay of
  \(N\), and the dashed line indicates the tangent. (b) is the error rate curve zoomed-in from 180 to 200, and the dashed line represents the fitting.}
\label{fig:figure1label1}
\end{figure}

The Fig.\ref{fig:figure1label1} shown the different polynomial of \(N\) to training the ResNet on the CIFAR-10 dataset. \textit{We found that the ROC of the errors will vary with the
ROC of LR, has positive correlation, especially at the end stage. So we suppose
increase the ROC of the LR to increase of ROC of the errors, be equivalent to
improves the model's performance, at the end of training period, but how to
increase the ROC of the LR?}

In this paper, we purpose a method based on mathematical derivatives, name k-decay, which by impacting its
k-th order derivative, to increase the ROC of original LR schedule, at the end
stage. Denote the function of original LR
schedule by \(\eta(t)\) and its k-th order derivative function by
\(\eta^{k}(t)\), then \(\eta^{k}(t)\) is updated in the following way,

\begin{equation}
\label{eq:orgc2db203}
\boxed{\eta^{k}(t)'=\eta^{k}(t) + \Delta f^{k}(t),}
\end{equation}
where \(k \in \mathbb{N}\). The \(\Delta f^{k}(t)\) is the increment of
\(\eta^{k}(t)\), which controls the ROC of \(\eta(t)\) in the k-th order. Eventually,
the solution of \(\eta^{k}(t)'\) is the \(\eta(t)'\) for new LR schedule.
Obviously, the solution \(\eta_o(k, t)\) must contain \(k\), so named k-decay.

\begin{figure}[!t]
  \centering
  \centerline{\includegraphics[width=\linewidth]{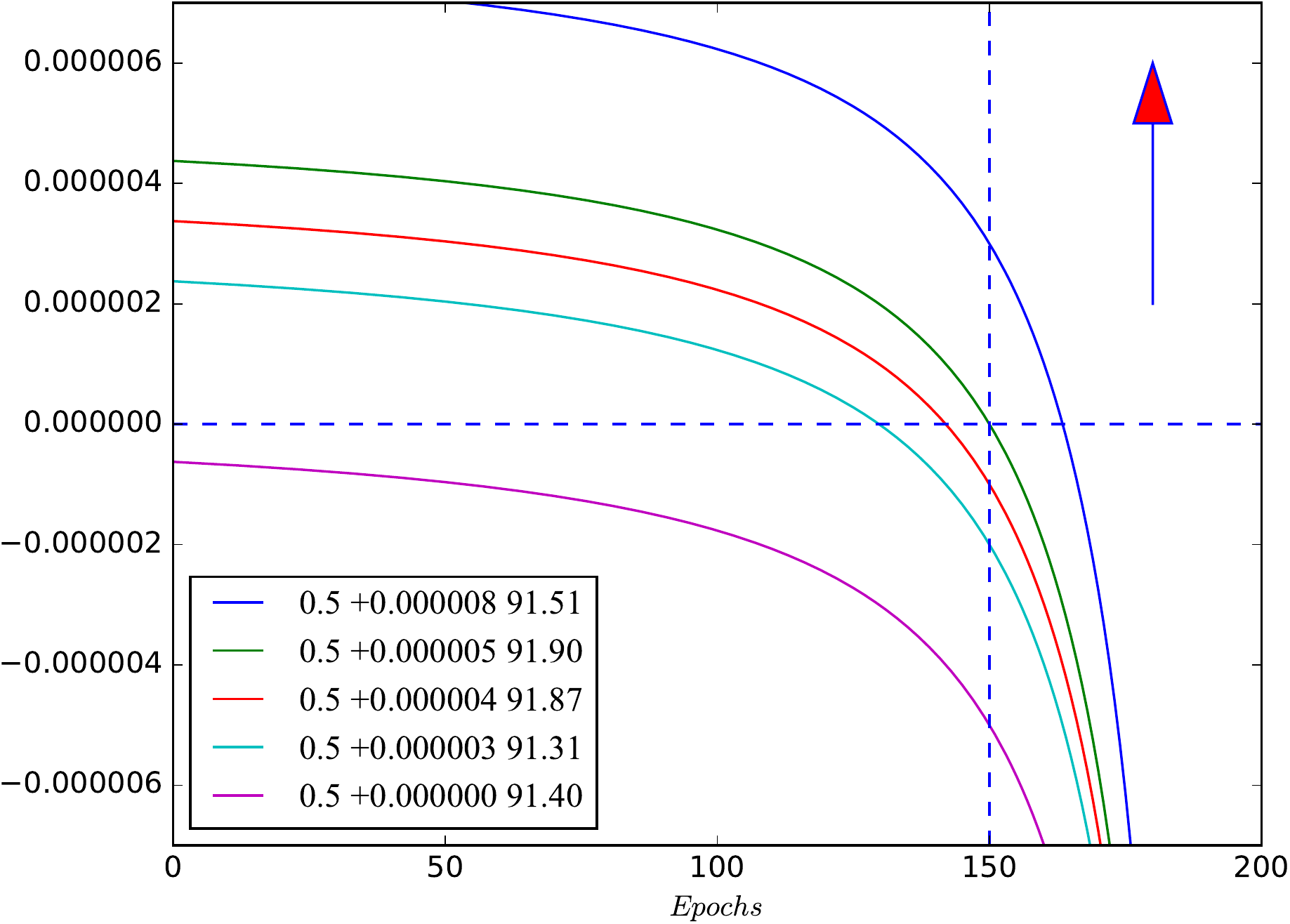}}
  \caption{The derivative function of polynomial function at \(N=0.5\) with
    different increment \(\alpha_0\). As \(\alpha_0\) increases, the accuracy
    also increases within the appropriate range.}
  \label{fig:figure1label2}
\end{figure}

In order to preliminarily verify the effectiveness of this method, we simplified
the equation through the following steps and tested its effect. \textbf{STEP1:}
Find its k-th derivative \(\eta^{k}(t)\). \textbf{STEP2:} Add an increment
\(\alpha_{0}\): \(\eta^{k}(t)+\alpha_{0}\). \textbf{STEP3:} Solve this used for
training. The Fig.\ref{fig:figure1label2} show the 1-th derivative graph of the
polynomial LR schedule at \(N=0.5\), we gradually increase \(\alpha_{0}\), the accuracy is indeed improved by
\(0.5\%\) at \(\alpha_{0}=0.000005\). Also we have done many other examples and
\(k\)-th derivative, have the same effect.

In Sec.\ref{sec:kdecay}, we
derivative a special solution \(\eta_o(k, t)\) as a standard procedure based on polynomial LR schedule and above steps. It should be pointed
out that there is no particular preference for our choice of \(\eta(t)\) like
polynomial LR schedule, we can still choose other LR shcedule for special
solutions of k-decay. Then we uesed the solution as an additional item to other LR schedule, the
performance of these LRs has been significantly improved. In summary, our kdecay method is proposed with wide applicability, which can
obtain different forms of variants when applied to different LRs, and can
greatly improve the performance of the model.

\section{Related Work}
\label{sec:orge7e4fb2}
Recent years have seen many optimization studies based on the Learning Rate. On
the one hand, in the better LR schedule role, the global LR change with time in the optimal algorithm.
Such examples include the multi-step decay, polynomial decay, Stochastic Gradient
Descent with Restarts (SGDR)~\cite{loshchilov2016sgdr}, Cyclical Learning Rates (CLR)~\cite{conf/wacv/Smith17},
and Hyperbolic-Tangent Decay (HTD)~\cite{conf/wacv/HsuehLW19}. The SGDR uses the
cosine decay, combined with periodic function. The CLR is a periodic
function, which uses the maximum and minimum values as a period. The HTD uses the tanh
function to construct a new LR schedule. A recent work
AutoLRS~\cite{jin2021autolrs} purpose a automatic LR schedule by bayesian. All these schedules can improve the
performance of the model of interest. But, compared with our method, these methods have more hyperparameter
settings, and the training process is extremely fluctuating. Our method has only
one hyperparameter \(k\), and the performance is positively correlated with
\(k\). Our method has wide applicability, and different forms
of variants will be obtained when applied to different LR schedules.

On the other hand, initialization parameters are changed based on the history
of the gradient. For example, some adaptive learning rates algorithm: RMSprob~\cite{ruder2016overview}, AdaDelta~\cite{journals/corr/abs-1212-5701}, Adam~\cite{kingma2014adam}. These methods improve the stochastic gradient descent based
on the momentum. It can accelerate the convergence and reduce the number of steps of the
LR. However, such a method is not contradiction with the LR schedule, with better performance when they are combined.

\section{k-decay For Learning Rate schedule} \label{sec:kdecay}
In the section, we derivation a special solution of k-decay based on the
liner polynomial function and then generalize to other LR shcedule. The liner polynomial function is
\begin{equation}
\label{eq:orgde34395}
\eta(t)=(\eta_{0}-\eta_{e})(1- \frac{t}{T})+\eta_{e}.
\end{equation}

For simplicity in calculation, we consider here the \(\Delta f^{k}(t)\) in the k-decay equation (eq.(\ref{eq:orgc2db203})) being constant with time:
\begin{equation}
\label{eq:org884107b}
    \eta^{k}(t)' = \eta^{k}(t) + \Delta f^{k}_0,
\end{equation}
So the new liner polynomial function can be
\begin{equation}
\label{eq:org5f46a8c}
    \eta(t)' = (\eta_{0}-\eta_{e})(1-\frac{t}{T})+\eta_{e} + \eta_{o}(k, t),
\end{equation}
\(\eta_{o}(t)\) is additional terms raised by the $\Delta f^{k}_0.$ 
And the boundary condition is
\begin{equation}
\left\{\begin{matrix}
\eta(t+1)' \leq \eta(t)'
\\
\eta(0)' = \eta_0, \eta(T)' = \eta_e
\end{matrix}\right.
\end{equation}
Series expansion of \(\eta_{o}(k, t)\) leads to,
$$\eta_{o}(k,t) = a_{k}t^{k}+...+ a_{1}t+a_{0},$$
Without loss of generality, let \(a_{k-1}=0,...,a_{0}=0\), we have
\begin{equation}
\label{eq:orgbf53052}
   \eta(t)' = (\eta_{0}-\eta_{e})(1-\frac{t}{T})+ \eta_{e} + a_{k}t^{k}+ a_{1}t.
\end{equation}
Then \(k\) order of eq. (\ref{eq:orgbf53052}) is given by,
\begin{equation}
\label{eq:orgd504745}
\eta^{k}(t)'=\eta^{k}(t)+k!a_{k} =0+k!a_{k}=k!a_{k},
\end{equation}
so 
\begin{equation}
\label{eq:orgd504745}
\Delta f^{k}_0 = k!a_{k}.
\end{equation}
We find the $k$-th order of the function \(f(t) =
(\eta_{0}-\eta_{e})(1-\frac{t}{T})^n+\eta_{e}\) is constant with time at \(n = k\):
$$f^{k}(t)=(\eta_{0}-\eta_{e}){k!}(-\frac{1}{T})^{k}t^{n-k}=(\eta_{0}-\eta_{e})k!(-\frac{1}{T})^{k}.$$
It can be used as a special solution of the eq.(\ref{eq:orgbf53052}). Let \(\Delta f^{k}_0=f^{k}(t)\), we have
$$a_{k}= \pm (\eta_{0}-\eta_{e})\frac{1}{T^k}.$$
Consider the boundary conditions, substitute into the eq.(\ref{eq:orgbf53052}) , and then simplified to
\begin{equation}
\label{eq:org421ac02}
\eta_o(k,t) = (\eta_{0}-\eta_{e})(\frac{t^k}{T_0^k}-\frac{t}{T_0}).
\end{equation}

The special solution can be uesd as an additional
term, added to other LR schedule, bulid a new LR schedule of the k-decay. In the
LR schedule, we control the increment \(\Delta f^{k}(t)\) by \(k\). In fact, as
you can see in Fig.\ref{fig:figure1label2}, adjusting the incremental
\(\alpha_{0}\) is equal to \(k\). Specialy, the original function is a special case at \(k =1\).

The LR schedule of k-decay is
\begin{equation}
  \begin{aligned}
  \label{eq:org421ac02}
  \begin{split}
  \eta(t)' &= \eta(t) + \eta_{o}(k,t)\\
  &= \eta(t) +(\eta_{0}-\eta_{e})(\frac{t^k}{T_0^k}-\frac{t}{T_0}),
\end{split}
\end{aligned}
\end{equation}
where \(\eta_{0}(t)\) is our additional item, \(\eta(t)\) is the orignal LR schedule, \(t\) is the current time, \(T_0\) is the total time.


For instance, the polynomial (POL) of k-decay is,
\begin{equation}
\label{eq:org444e460}
\eta(t)'=(\eta_{0}-\eta_{e})(1-\frac{t}{T_0})^N + \eta_{e} + \eta_{o}(k,t).
\end{equation}

The cosine (COS) of k-decay is:
\begin{equation}
\label{eq:org9e8c565}
\eta(t)' = \frac{1}{2}(\eta_0-\eta_e)(1+\cos(\frac{t}{T_0}\pi)) + \eta_e +  \eta_{o}(k,t).
\end{equation}

The exp (EXP) of k-decay is:
\begin{equation}
\label{eq:org840651e}
\eta(t)' = (\eta_0-\eta_e)\exp(-\frac{t}{T_0}) + \eta_{o}(k,t).
\end{equation}

\begin{figure}[!t]
\centering
\centerline{\includegraphics[width=0.5\textwidth]{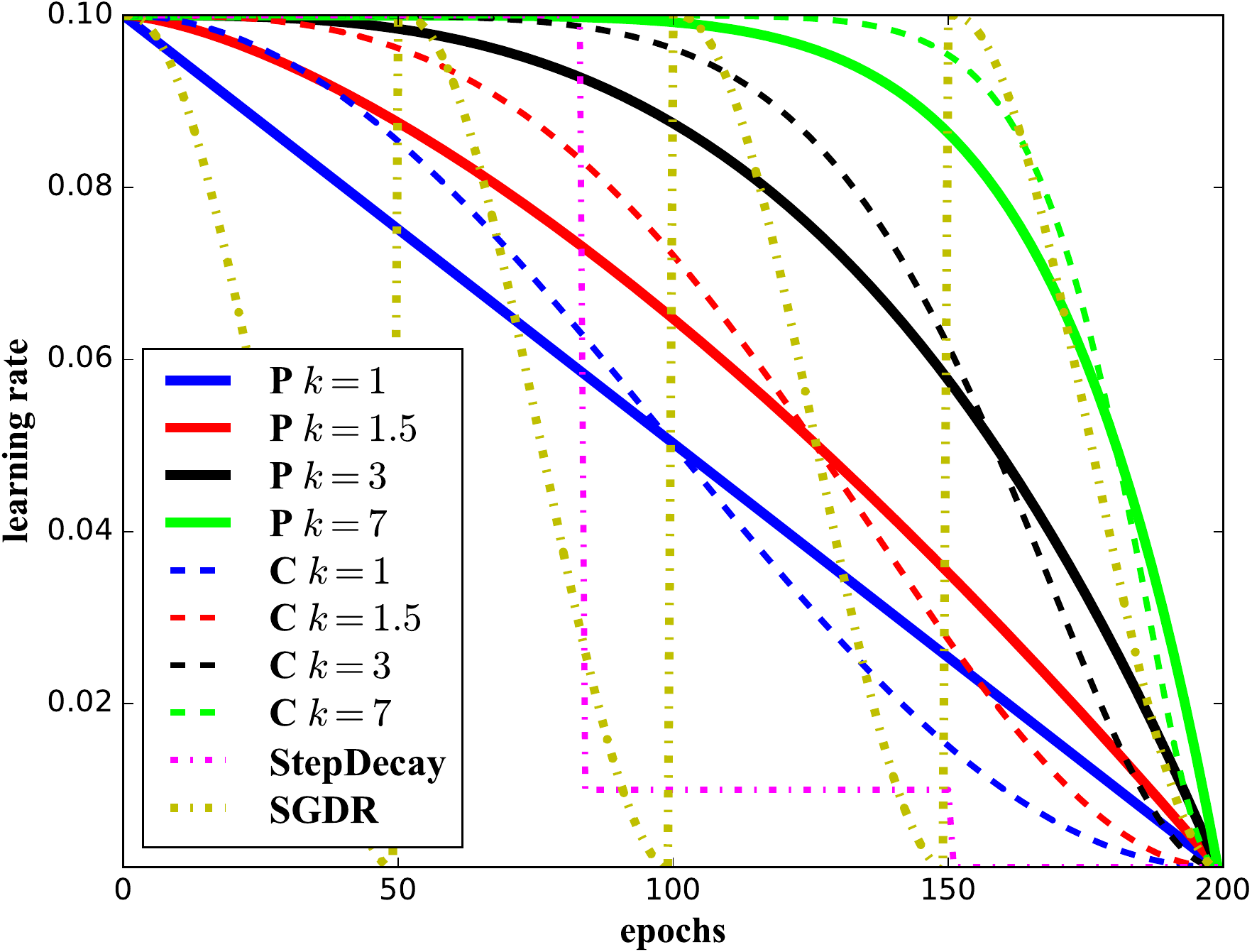}}
\caption{The learning rate curve of k-decay at different $k$ with the linear
  polynomial decay \textbf{(P)} and cosine decay \textbf{(C)}. The $\eta_{0}$ is
  0.1, $\eta_{e}$ is 0.001. According to the Eq.(\ref{eq:org421ac02}) and
  Eq.(\ref{eq:org9e8c565}). And our baseline model use the step decay and SGDR ~\cite{conf/wacv/HsuehLW19}.}
\label{fig:figure1label4}
\end{figure}
Fig. \ref{fig:figure1label4} shows the difference between the original LR schedule
and the new LR schedule of the k-decay. The ROC of the LR in the late training
period increases with the
increase of \(k\) in the new LR schedule. Applying the k-decay factors to different
functions is equivalent to changing the ROC of the LR with \(k\).

\section{Experiments}
\label{sec:org9946392}

\textbf{Datasets} We choose the accuracy of classification tasks on CIFAR~\cite{Krizhevsky09} and
ImageNet~\cite{imagenet_cvpr09} as the standard to measure performance. The CIFAR datasets, including 10 categories of CIFAR-10 and 100 categories of CIFAR-100, which are respectively
composed of 50,000 training sets and 10,000 test sets. The ImageNet data set
includes 1000 classified. Here use the ILSVRC2012 classification
data set, which has 1.28 million training pictures and 50k validation sets. 

\textbf{Baseline} Usually for convenience, we will choose Step Decay, EXP, COS
and POL as the learning rate schedule, and of course the latest SOAT algorithms
such as SGDR~\cite{conf/wacv/HsuehLW19} and CLR~\cite{conf/wacv/Smith17}. We will k-decay item is added to the commonly used LR
schedule EXP. COS and POL, to compare the performance difference with original LR schedule, to illustrate
the versatility of our method, is also compared with SDGR and CLR to illustrate the superiority of our method.

\textbf{Implementation} The implementation of the Wide ResNet-28-10 for CIFAR
and  ~\cite{he2016deep} the ResNet-50 for ImageNet~\cite{zagoruyko2016wide} is the same as the
original paper . The optimizers used SGD with momentum, and momentum of 0.9. The Wide ResNet-28-10 training 200
epoch, the ResNet-50 training 90 epoch. We used different LR schedule for optimizers,
where \(\eta_0\) of 0.1, and \(\eta_e\) of 0.001. The
\(t\) is set by the batches number, makes the change of LR more
continuous.  

\label{sec:org8fec2ee}

\renewcommand\arraystretch{1.0}
\begin{table}
  \begin{center}
    \resizebox{\linewidth}{!}{
      \begin{tabular}{lll}
   \Xhline{2pt}

  Method & CIFAR-10 & CIFAR-100 \\
  \hline  \hline 
  CLR~\cite{conf/wacv/Smith17} & 4.93  & 21.58 \\
  StepDecay~\cite{zagoruyko2016wide} & 4.17 & 20.50\\
  SGDR~\cite{conf/wacv/HsuehLW19} & 4.03 & 19.58 \\
  
  \hline
  EXP  & 4.11 & 20.40 \\
  EXP with k-decay (ours) &\textbf{4.03\textsubscript{\(2.0\)}} $\uparrow$ 0.08 & \textbf{19.05\textsubscript{\(2.0\)}} $\uparrow$ 1.35 \\
  \hline
  COS  &  3.68 &  18.68 \\
  COS with k-decay (ours) &\textbf{3.82\textsubscript{\(1.5\)}} $\downarrow$ 0.14 & \textbf{18.44\textsubscript{\(5.0\)}} $\uparrow$ 0.24 \\
  \hline
  POL  & 3.95 & 19.42 \\
        POL with k-decay (ours) &\textbf{3.59\textsubscript{\(1.5\)}} $\uparrow$ 0.36 & \textbf{18.43\textsubscript{\(1.5\)}} $\uparrow$ 0.99 \\
        \Xhline{2pt}
  
      \end{tabular}
     }
\end{center}
\caption{\label{tab:orga1d02d9}Testing error (\%) on CIFAR-10 and CIFAR-100
  datasets on Wide ResNet-28-10. The overall best results are bold. For instance in our results denoted by \([5.26_{2.0}]\), \(5.26\)
  means the errors (\%), subscript \(2.0\) means \(k=2.0\) on k-decay.}
\end{table}


\begin{figure}[!ht]
  \centering
  \centerline{\includegraphics[width=\linewidth]{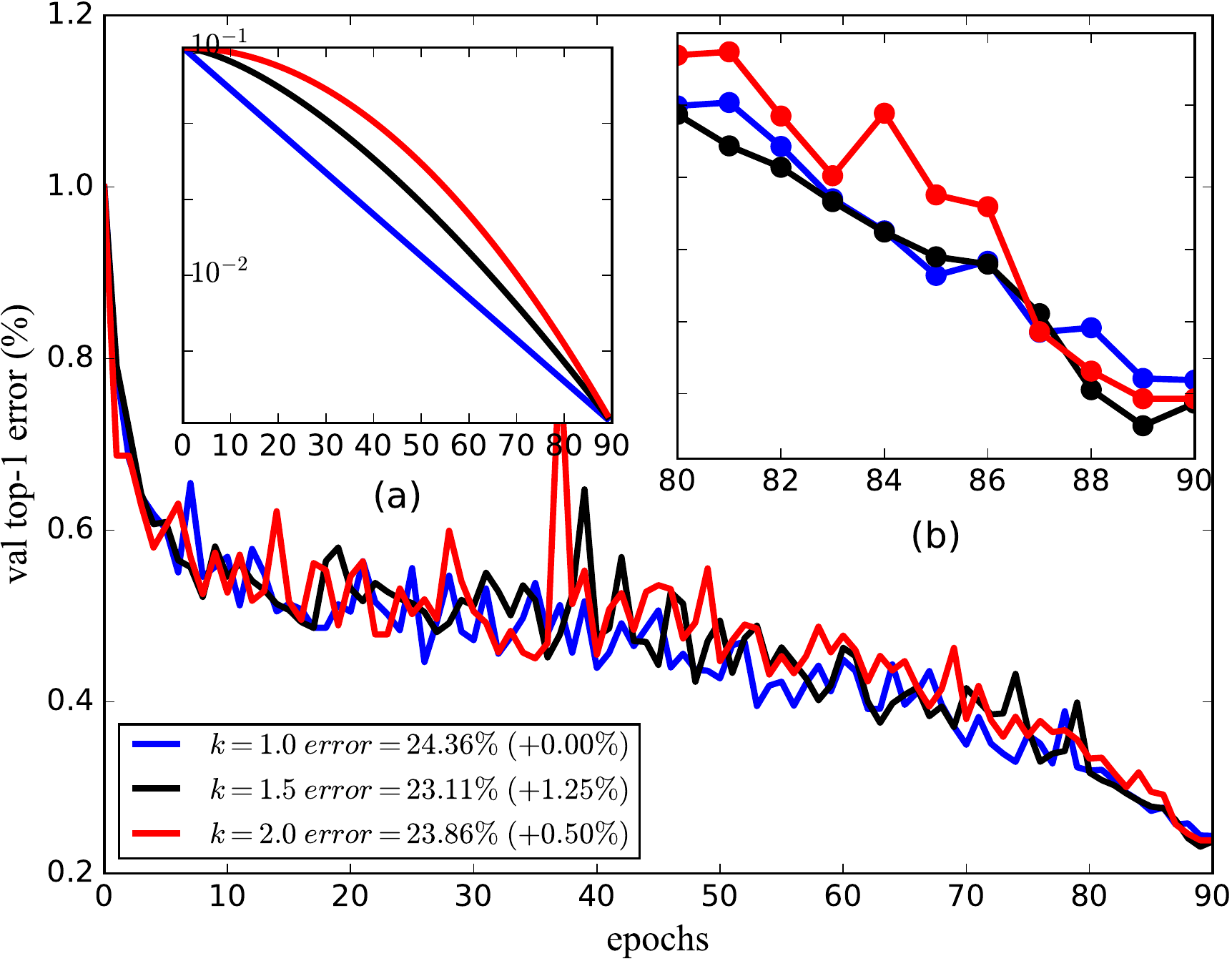}}
  \caption{Error rates (\%, single-crop testing) on ImageNet validation with model ResNet-50. (a) is the learning rate curve of the new liner polynomial decay function of different $k$. (b) is the error rate curve zoomed in from 80 to 90.}
  \label{fig:figure1label12}
\end{figure}
\textbf{Results on CIFAR Datasets} The main results of tests on CIFAR-10 and CIFAR-100 are shown
in Table \ref{tab:orga1d02d9}. The best state-of-the-art results are
marked in bold. Compared with the  Multi-Step Decay baseline method, the
accuracy of POL with k-decay is improved by 0.58\% on the CIFAR-10 and
2.07\% on the CIFAR-100. Comparing with the recent work SGDR, our results
(POL with k-decay) are better
than 0.44\% on CIFAR-10 and 1.15\% on CIFAR-100. Comparing with the CLR, our results
(POL with k-decay) are better
than 0.44\% on CIFAR-10 and 1.15\% on CIFAR-100. Compared to themselves (EXP,
COS and POL), our method is better than the original function, can improve up to
1.34\% on CIFAR-10 and 3.15\% on CIFAR-100. But there is a performance drop in
COS with k-decay on CIFAR-10, we think this because the k-decay term derived on
POL does not apply to COS, the derivative of POL is a monotonic function, COS is
not, we can also see that its performance improvement on POL is more obvious.

\label{sec:org8fec2ssee}

\renewcommand\arraystretch{1.0}
\begin{table}
\begin{center}
\begin{tabular}{ll}
\Xhline{2pt}
Method & Top-1 err. \\
  \hline  \hline 

  StepDecay ~\cite{chen2021xvolution} & 24.30 \\
  SGDR ~\cite{conf/wacv/HsuehLW19} &24.14   \\
  AutoLRS ~\cite{jin2021autolrs} &24.07   \\
  CLR ~\cite{conf/wacv/Smith17} &  24.06 \\
  
  \hline
  POL  & 24.36  \\
  POL with k-decay (ours) &\textbf{23.11\textsubscript{\(1.5\)}}$\uparrow$ 1.25  \\
  \Xhline{2pt}

\end{tabular}
\end{center}
\caption{\label{tab:orga1d02d9}Top-1 error (\%) on ImageNet datasets with ResNet-50. The overall best results are bold. For instance in our results denoted by \([5.26_{2.0}]\), \(5.26\)
  means the errors (\%), subscript \(2.0\) means \(k=2.0\) on k-decay.}
\end{table}

\textbf{Results on ImageNet Datasets} In Fig.  \ref{fig:figure1label12}, we
 employ the POL with k-decay to train the ResNet-50
model ~\cite{he2016deep} on the ImageNet datasets. The result indicates that the model
accuracy is improved by 1.25\%, when we set $k=1.5$ than original method (\(k=1\)). And training
the ResNet-50 with Step Decay is 24.3\% ~\cite{chen2021xvolution}, and ours
method is better than 1.19\%. In Table.\ref{tab:orga1d02d9} We also compared
with recent works SGDR ,AutoLRS and CLR,
the accuracy improved by 1.03\% ,0.96\% and 0.95\%.

\section{Discussion}
\begin{figure}[!t]
\centering
\centerline{\includegraphics[width=\linewidth]{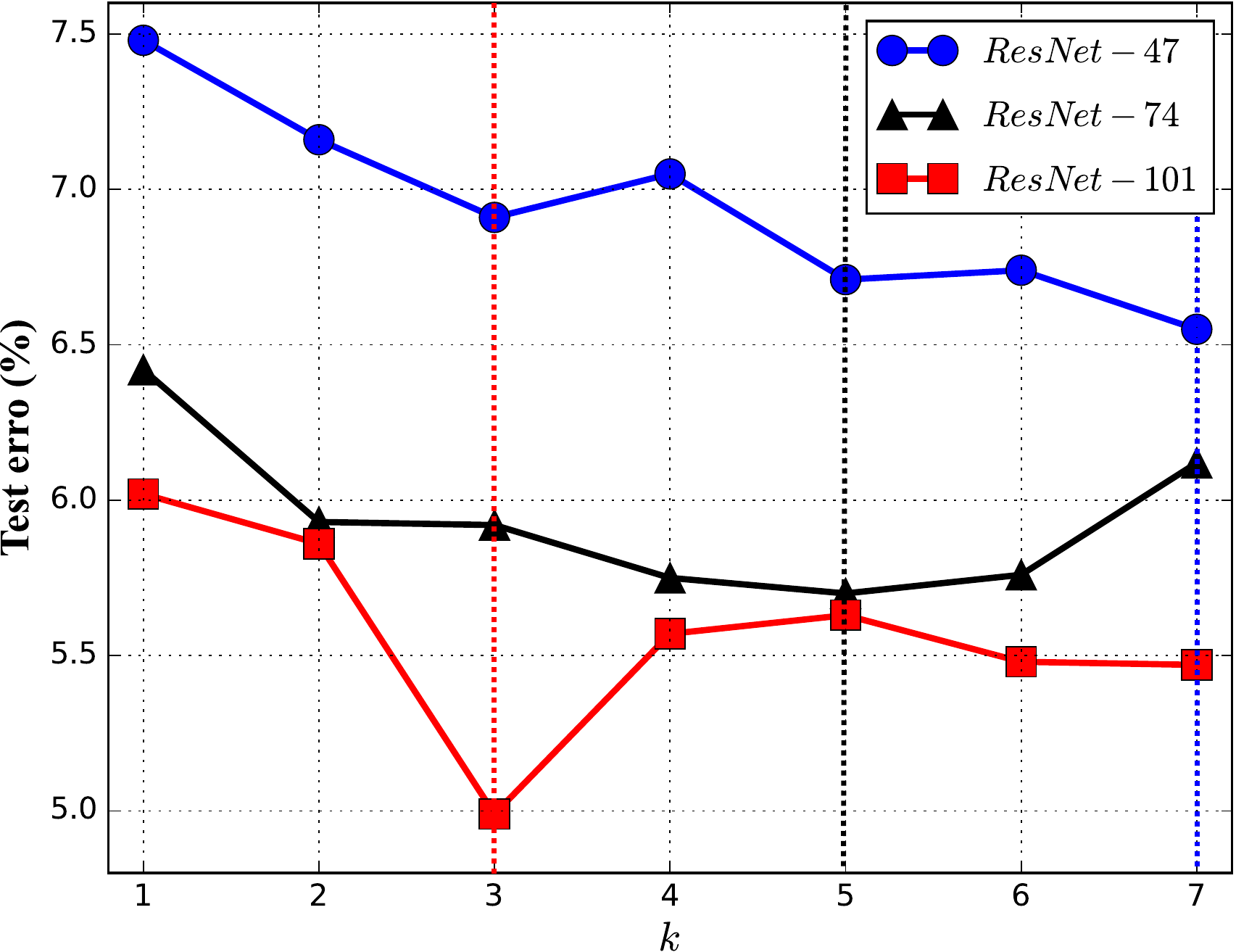}}
\caption{The test errors rate obtained by training ResNet of different depths (ResNet-47, ResNet-74, ResNet-101) with continuous $k$ values on the CIFAR10 data set. In a specific range, the error rate decreases as the value of $k$ increases.}
\label{fig:figure1label7}
\end{figure}

\textbf{Impact of \(k\) on Performance} Fig.\ref{fig:figure1label7} shows the
relation between error rate and the \(k\) value on residual neural networks with
different depths (ResNet-47, ResNet-74, ResNet-101). The test accuracy increases
with the increase of the hyperparameter $k$  (starting from $k = 1$).
However, when \(k>k_v\), the accuracy starts to drop. In ResNet-101, the \(k_v\)
is 3. In ResNet-47, \(k_v\) should be 7. We can found that the threshold value
\(k_v\) of the model will be decreasing with the increase of the model's depth.
It reflects that deeper models are more sensitive to the LR's ROC than shallow
ones. The threshold is different for different models and datasets. According to
our experiments, the recommended \(k \) is 1.5.

\begin{figure*}[!t]
\centering
\begin{minipage}[t]{0.48\linewidth}
\centerline{\includegraphics[width=8.2cm]{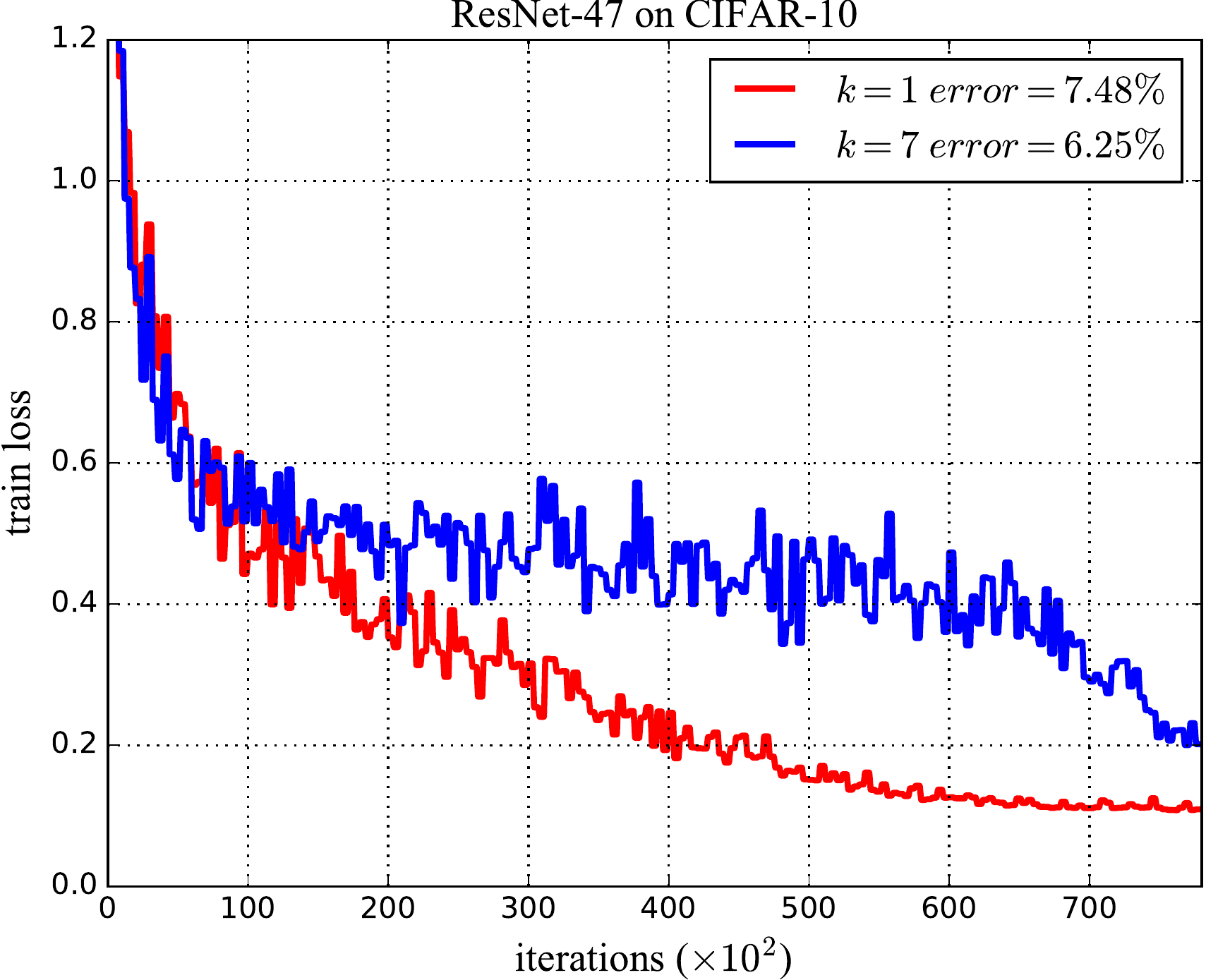}}
\end{minipage}
\begin{minipage}[t]{0.48\linewidth}
\centerline{\includegraphics[width=8.2cm]{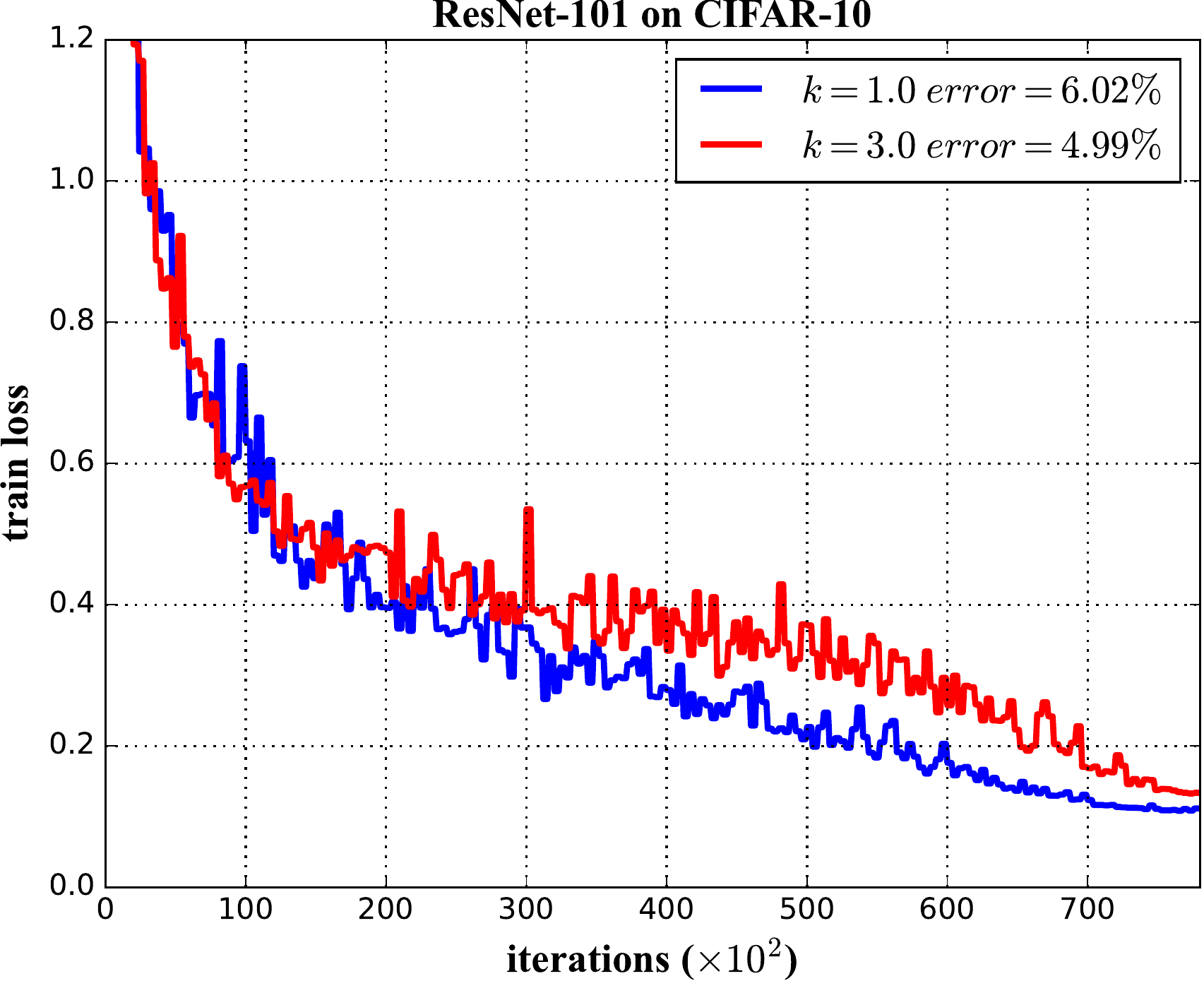}}
\end{minipage}

\begin{minipage}[t]{0.48\linewidth}
\centerline{\includegraphics[width=8.2cm]{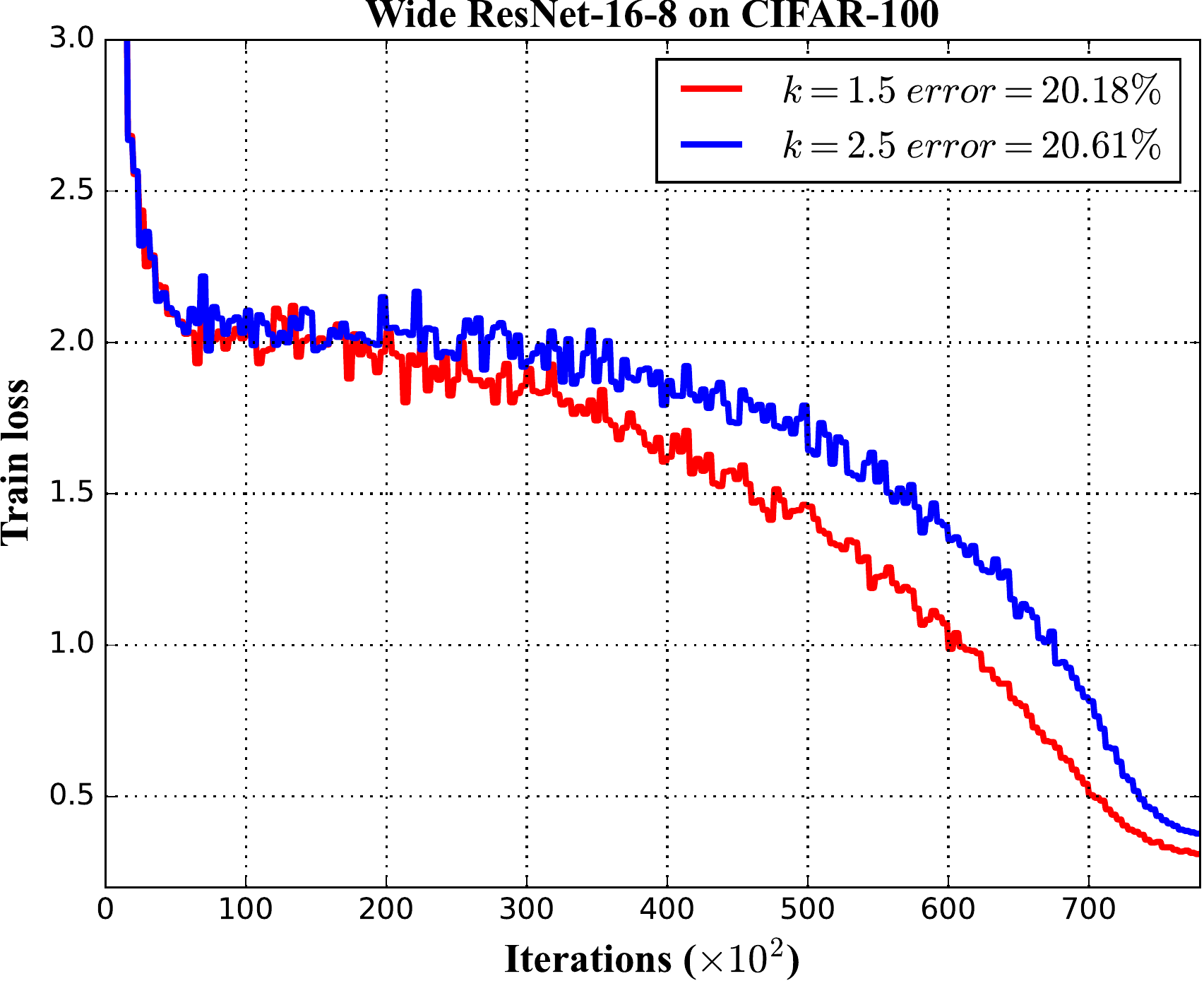}}
\end{minipage}
\begin{minipage}[t]{0.48\linewidth}
\centerline{\includegraphics[width=8.2cm]{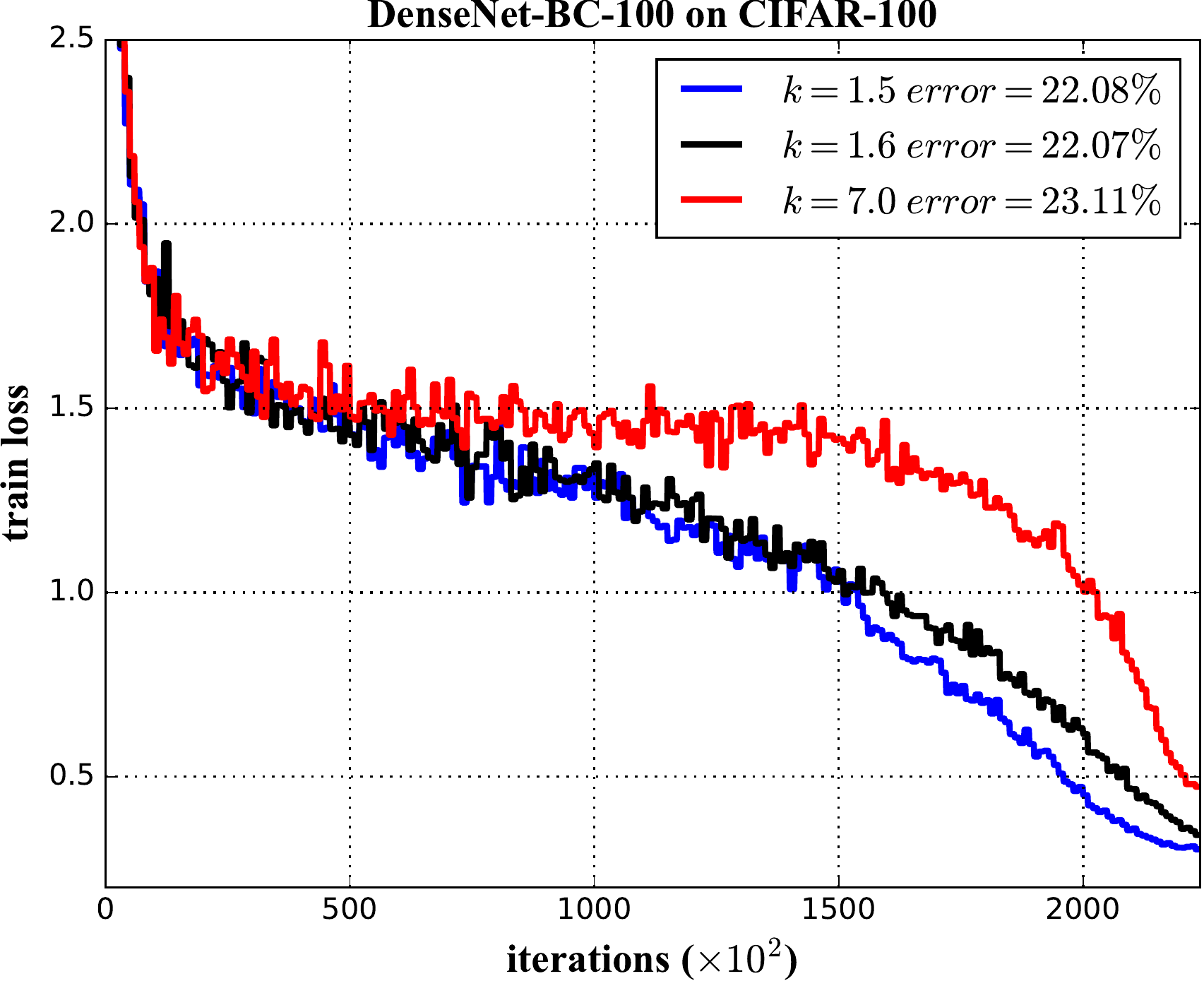}}
\end{minipage}

\begin{minipage}[t]{0.48\linewidth}
\centerline{\includegraphics[width=8.2cm]{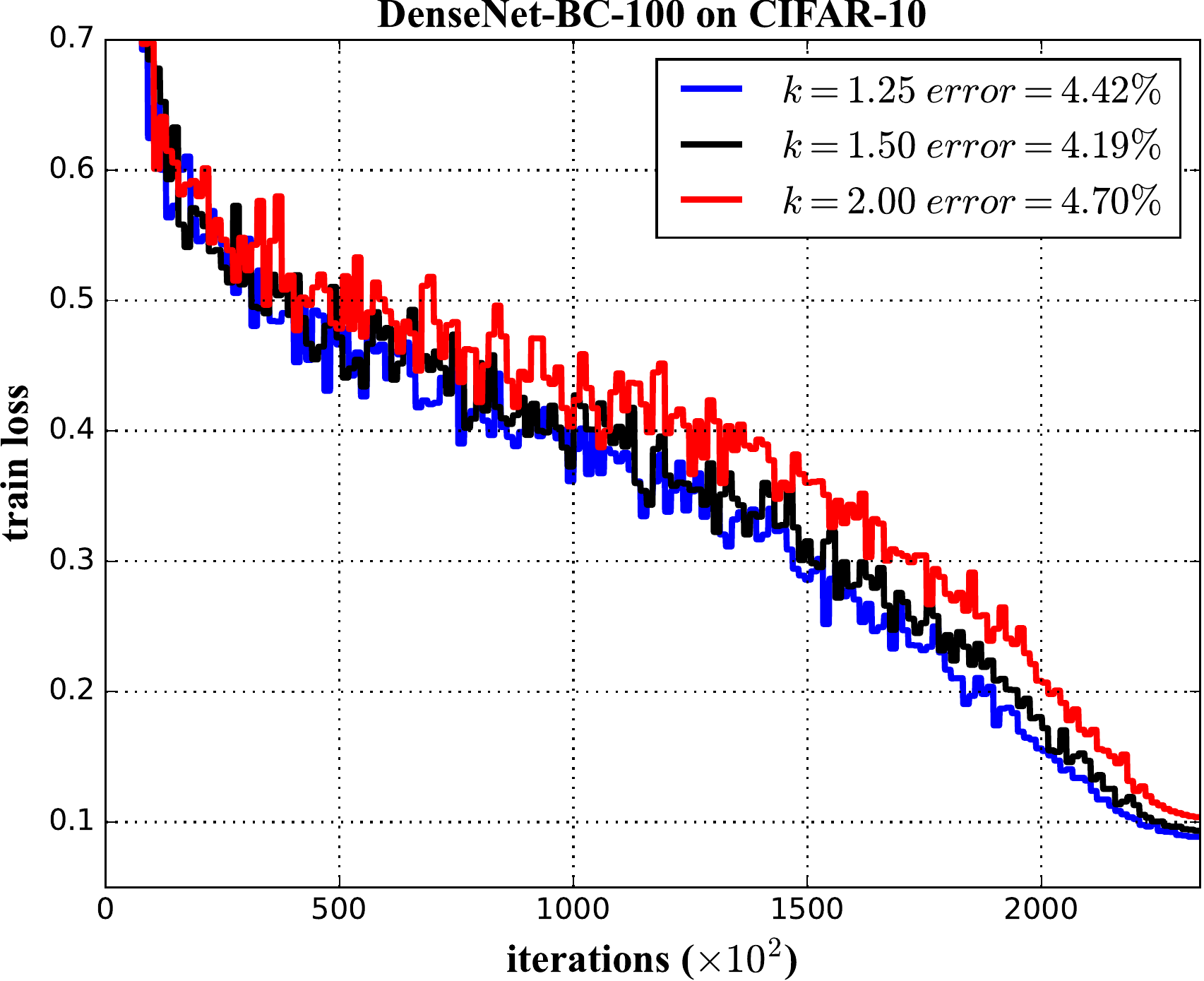}}
\end{minipage}
\begin{minipage}[t]{0.48\linewidth}
\centerline{\includegraphics[width=8.2cm]{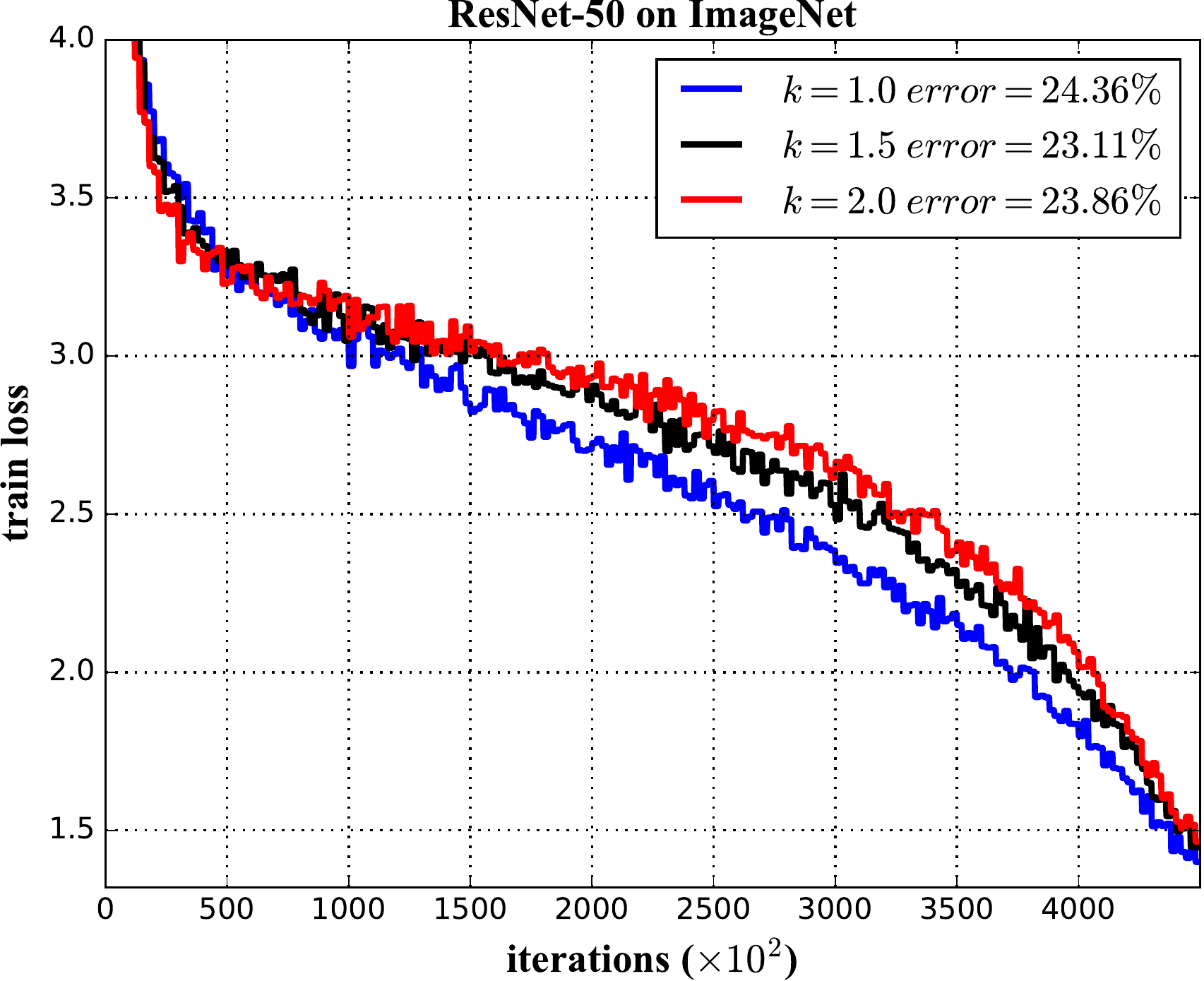}}
\end{minipage}

\caption{The loss curves of the ResNet, Wide ResNet, DenseNet during the training period. The values of \(k\) in the red line are greater than the values of $k$ in the blue line. The red line always tops the blue one in the loss curves. The red line is steeper than the blue one at the later stage because the ROC of the LR increases with $k$.}
\label{fig:figure1label8}
\end{figure*}

\textbf{Impact of \(k\) on Training} According to Fig. \ref{fig:figure1label8},
we found that $k$ can effectively affect the training process. The loss curve of
large $k$ is always above the loss curve of small $k$. And also the ROC of the
loss with an increase of $k$ at the later stage. Due to the ROC of the LR with
the increase of $k$ on the new function. Larger \(k\) makes that the ROC of the
LR at the early and intermediate stages will be smaller, which makes the ROC of
the loss smaller, lead to the speed of the convergence is slowing down,
resulting in a minor loss. At the later stage, the ROC of the LR will be giant,
lead to the rapid convergence of the loss function to compensate for the loss
initially. It is why the performance increase with \(k\). However,
when $k$ is too big, the intermediate stage's loss is too low, the speed of the
convergence cannot make up at the later stage. It is the reason the performance
degradation at \(k\) value is greater than \(k_v\). 

\section{Conclusion}
\label{sec:orgdb3c189}

This paper proposes a new method for the learning rate schedule by using the
$k$-th order derivatives to obtain a new function. A specific solution of
k-decay \(\eta_o(k, t)\) is derived, which that can be widely used in other LR schedules. In the k-decay method, we introduce a hyper-parameter \(k\)
to control the LR's ROC in the new function, which enriches the LR schedule functions. The experiments show how the accuracy improvement changes with the increase of
\(k\). It is proved that
the k-decay method is effective and easier to use.
This article studies the situation where time is not involved and finds a
particular solution. For the situation when time is considered,
there may even be other better forms of particular solutions for different LR schedules.

\section{Acknowledgement}

We gratefully acknowledge the fruitful discussions with Yueying Zhu. Thanks to
Shengtai Li of the Nuclear Science Computing Center at
Central China Normal University, Wuhan, China. This work was supported in part
by the National Natural Science Foundation of China (Grant Nos. 61873104, 11505071, 11747135,
and 11905163), the Programme of Introducing Talents of Discipline to
Universities under Grant No. B08033, the Fundamental Research Funds for the
Central Universities (Grant No. KJ02072016-0170, CCNU, CCNU19QN029,
CCNU19ZN012), and the China Postdoctoral Science Foundation (Grant No. 3020501003).

{\small
  \bibliographystyle{ieee_fullname}
  \bibliography{find_lr_ref.bib}
}

\end{document}